%% file: root.tex
\documentclass[letterpaper, 10 pt, conference]{ieeeconf}  % Comment this line out if you need a4paper
\IEEEoverridecommandlockouts                              % to use the \thanks command
\title{\LARGE \bf
\nas: N-step computation of All Solutions to the footstep planning problem %, or 31 ways to climb a staircase
}

\author{
   Jiayi Wang$^{1*}$,  Saeid Samadi$^{1*}$, Hefan Wang$^{1*}$, Pierre Fernbach$^{2}$, Olivier Stasse$^{3,4}$, Sethu Vijayakumar$^{1}$ \\ and Steve Tonneau$^{1}$\textsuperscript{\textdagger}%
    \thanks{$^{*}$Joint first authors. \textsuperscript{\textdagger}email: {\tt\small stonneau@ed.ac.uk}}
    \thanks{$^{1}$University of Edinburgh, School of Informatics, UK}%
    \thanks{$^{2}$TOWARD S.A.S., Toulouse, France}%    
    \thanks{$^{3}$LAAS-CNRS, Université de Toulouse, Toulouse, 31400, France}% 
    \thanks{$^{4}$Artificial and Natural Intelligence Toulouse Institute, Toulouse, France}%
    \thanks{This work was supported by a Tata Consulting Services grant, }
    \thanks{Dynamograde (ANR-21-LCV3-0002), ROBO-TEX 2.0  }
    \thanks{{(ROBOTEX ANR-10-EQPX-44-01 and TIRREX-ANR-21-ESRE-0015),}}
    \thanks{the JST Moonshot R\&D Grant No. JPMJMS2031, }
    \thanks{ANITI (ANR-19-P3IA-0004) and The Alan Turing Institute.}
}

\DeclareTextSymbolDefault{\dh}{T1}
\usepackage{hyperref}
\usepackage{graphicx}
\usepackage{amsmath} % assumes amsmath package installed
\usepackage{amssymb}  % assumes amsmath package installed
\usepackage{bm}  % assumes amsmath package installed
\usepackage{tikz}
\usetikzlibrary{shapes.geometric, arrows}
\usepackage{pgfplots}
\usepgfplotslibrary{fillbetween}
\usepackage{subcaption}
\usepackage{algorithm}
\usepackage[noend]{algpseudocode}
\usepackage{xcolor}

\pgfplotsset{compat=1.18}

\tikzstyle{bloc} = [rectangle, rounded corners, minimum width=2.5cm, minimum height=3cm, text centered, draw=black, fill=yellow!10, align=center]
\tikzstyle{decision} = [diamond, minimum width=2cm, minimum height=1cm, text centered, draw=black, fill=lime!10]
\tikzstyle{arrow} = [thick,->,>=stealth]
\tikzstyle{inout} = [anchor=south, align=center]
\input{math_commands}

\begin{document}

\let\@oldmaketitle\@maketitle% Store \@maketitle
\renewcommand{\@maketitle}{\@oldmaketitle% Update \@maketitle to insert...
\centering
  \includegraphics[width=1\linewidth]
    {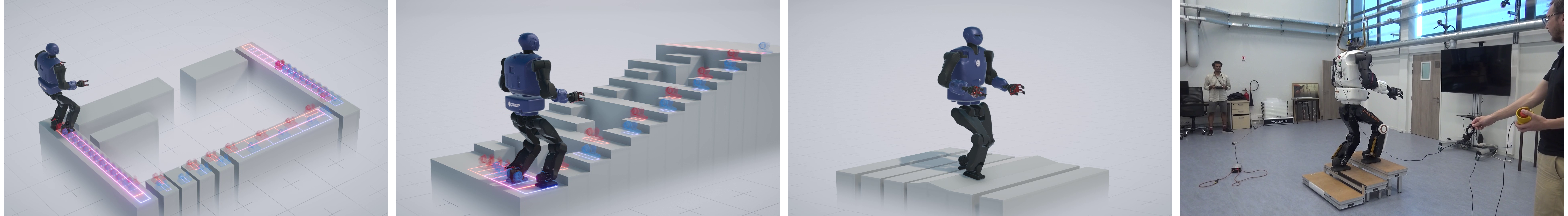}}% ... an image
\label{fig:teaser}

\maketitle
\thispagestyle{empty}
\pagestyle{empty}

%%%%%%%%%%%%%%%%%%%%%%%%%%%%%%%%%%%%%%%%%%%%%%%%%%%%%%%%%%%%%%%%%%%%%%%%%%%%%%%%
% ABSTRACT
%%%%%%%%%%%%%%%%%%%%%%%%%%%%%%%%%%%%%%%%%%%%%%%%%%%%%%%%%%%%%%%%%%%%%%%%%%%%%%%%
\begin{abstract}

How many ways are there to climb a staircase in a given number of steps? Infinitely many, if we focus on the continuous aspect of the problem. A finite, possibly large number if we consider the discrete aspect, \emph{i.e.} on which surface which effectors are going to step and in what order. We introduce \nas{}, an algorithm that considers both aspects simultaneously and computes \emph{all} the possible solutions to such a contact planning problem, under standard assumptions. To our knowledge \nas{} is the first algorithm to produce a globally optimal policy, efficiently queried in real time for planning the next footsteps of a humanoid robot. 

Our empirical results (in simulation and on the Talos platform) demonstrate that, despite the theoretical exponential complexity, optimisations reduce the practical complexity of \nas{} to a manageable bilinear form, maintaining completeness guarantees and enabling efficient GPU parallelisation. \nas{} is demonstrated in a variety of scenarios for the Talos robot, both in simulation and on the hardware platform. Future work will focus on further reducing computation times and extending the algorithm's applicability beyond gaited locomotion. Our video is available at \url{https://youtu.be/I5yFe0ez0sI}

\end{abstract}

%%%%%%%%%%%%%%%%%%%%%%%%%%%%%%%%%%%%%%%%%%%%%%%%%%%%%%%%%%%%%%%%%%%%%%%%%%%%%%%%
% INTRODUCTION
%%%%%%%%%%%%%%%%%%%%%%%%%%%%%%%%%%%%%%%%%%%%%%%%%%%%%%%%%%%%%%%%%%%%%%%%%%%%%%%%

\input{intro.tex}
\input{definitions}

\input{algo.tex}
\input{experiments}
\bibliographystyle{IEEEtran}{}
\bibliography{ref}
% \newline

\end{document}

%% file: math_commands.tex
\newcommand{\rfig}[1]{Fig.~\ref{#1}} 	
 	
\newcommand{\rfigtp}[2]{Fig.~\ref{#1}.#2}
\newcommand{\ralg}[1]{Algorithm~\ref{#1}} 	
\newcommand{\rstruct}[1]{struct~\ref{#1}} 	

\newcommand{\nas}[0]{NAS}
\newcommand{\astar}[0]{$A^*$}

\newcommand{\vc}[1]{\mathbf{#1}} 					% Vector symbol
\def\@onedot{\ifx\@let@token.\else.\null\fi\xspace}

\newcommand{\kdt}[0]{{k-d tree}} 	

\newcommand{\req}[1]{eq(\ref{#1})} 						% diag
\newcommand{\rseq}[1]{Section (\ref{#1})} 						% diag

\newcommand{\Expect}{{\rm I\kern-.3em E}}				% expectation

%% file: intro.tex
\section{Introduction}
In legged robotics, contact planning involves determining where an effector, such as a foot, should make contact with the environment to move. This process typically calculates a sequence of contacts to solve a particular locomotion problem instance, aiming for a \textbf{single} optimal solution. If this solution becomes unfeasible due to changes in the scene or drifts in state estimation during execution, a new plan must be computed. %In legged robotics, contact planning is the issue of identifying where an effector should create contacts with the environment to interact with it. The problem is often posed as the computation of a contact sequence that solves a given instance of the problem, possibly optimally. 
This paper aims at avoiding this costly re-computation by calculating \textbf{all possible solutions for a given instance}, specifically addressing bipedal locomotion.

Contact planning is crucial for any framework designed to generate legged motions, whether the goal is gaited locomotion (like footstep planning) or more complex loco-manipulation, which involves using all end-effectors in an arbitrary order. The challenge lies in handling non-linear geometric constraints (such as joint limits and collisions) and dynamic constraints, making it a high-dimensional, non-linear combinatorial problem.
%, whether the objective is gaited locomotion (this is the footstep planning problem) or more generally loco-manipulation, which involves using all effectors in an arbitrary order. 
%Planning contacts requires handling non-linear geometric (joint limits, collision) and dynamics constraints: this is a high-dimensional, non-linear combinatorics problem, which remains open to this day. 
While efficient heuristics exist for addressing simplified cases such as locomotion on flat ground~\cite{Kajita03}, the problem becomes more complicated when:

\begin{itemize}
    \item possible contact surfaces are not continuously connected (e.g., the steps of a staircase or stepping stones);
    \item Several steps need to be planned ahead to avoid dead-ends, causing an exponential increase in combinations.
\end{itemize}
%These issues worsen the combinatorics respectively in breadth ($m$ choices of surfaces are available for a single step) and depth (we need to plan the next $n$ steps), leading to an exponential increase in possible combinations ($m^n$).

%However, these approaches often rely on simplifying assumptions that are neither necessary nor sufficient.
Sampling-based approaches have been proposed to tackle the problem~\cite{tonneau2018efficient}, as well as other graph-search techniques such as \astar{} or Mixed-Integer Programming~\cite{chestnuttkuffner,deits2014footstep}. However, it is often required to re-plan contact plans online (at a 10Hz frequency \cite{risbourgsolo, corberes2024perceptive, Jenelten_2022_TAMOLS}) to account for execution and state estimation errors, especially in dynamic environments. The worst-case exponential complexity of these approaches then requires to significantly reduce the planning horizon $n$, that is the number of future steps that can be planned in advance. This is problematic in environments containing dead-ends as a short planning horizon does not allow to escape them.

Deep Reinforcement Learning (DRL) has shown promise in overcoming the need for simplifying assumptions, allowing robots to traverse challenging terrains robustly. However, so far DRL methods do not consider a long planning horizon. Preliminary research suggests that this horizon could be learned if combined with optimal control, but the generalisation is not trivial as these methods are supervised~\cite{wangvalue23, ravi2024efficientsearchlearningagile}. %Better delimiting the search space, even with over-approximations from graph-based methods, can enhance the efficiency of exploration and learning in DRL, facilitating faster convergence and better generalization to new situations.

%Deep Reinforcement Learning (DRL) approaches have recently demonstrated their efficiency in computing policies that enable legged robots to robustly traverse challenging terrains. DRL does not require simplifying assumptions and, despite their lack of theoretical guarantees, tends to generate robust motions. However, to date DRL is mostly relevant when a human user guides the motion with a joystick, meaning the planning horizon is managed externally.

%There is preliminary evidence that the planning horizon can also be learned~\cite{wangvalue23, silver2016mastering}, at the cost of exponentially time-consuming training. Better delimiting the search space for this problem, even using the over-approximations of graph-based approaches, would facilitate more efficient exploration and effective policy learning, enabling RL agents to achieve faster convergence and enhanced generalisation to new, unseen situations in real-world applications.

We propose to explicitly characterise all feasible solutions for a contact planning problem instance, up to a number $n$ of steps. This is achieved by observing that, under common assumptions, the set of feasible contact positions reachable in at-most $n$ steps can be computed recursively through the computation of Minkowski sums and polytope intersections similarly to the idea of backward reachability analysis in control theory~\cite{bansal2017hamilton,althoff2021set}. This effectively results in an optimal policy for computing minimum-step contact sequences given any state of the robot. \nas{} can also be implemented as a \astar{} for single-query planning, but we argue that outside of the theoretical interest, computing a globally optimal policy enables robust online replanning for disturbance adaption.  %\textcolor{blue}{The idea of our approach is similar to the backward reachability analysis studied in control thoery and hybrid systems~\cite{bansal2017hamilton,althoff2021set}, where the set of all initial states from which a system can reach a target set is determined by solving the Hamiltonian-Jacobi Equation. This technique has been widely applied in robotics for motion planning under uncertainty and safety verification~\cite{lengagne2011planning,liu2017provably,borquez2023hamilton}. In this paper, we focus on the contact planning problem and propose a similar approach that finds all feasible solutions of legged locomotion by backward propagating the reachability of contacts. }

%This policy is computed through the use of a novel Dynamic Programming algorithm, which takes as input the desired target state (expressed either as a point or a convex polygon) and either a desired depth $n$, or the current state of the robot. It outputs an acyclic oriented graph for which each node is not associated to a point, but rather to a convex polygon of possible positions for an effector. For a node at depth $i$, the graph guarantees that any point in the polygon allows to reach the target state in $i-1$ steps, and that this is the minimal number of steps required to reach the target. 

%Characterising the feasible space is not only practically beneficial but also crucial for advancing machine learning techniques. Most contact planning methods, including ours, rely on simplifying assumptions about the geometric and dynamics constraints of the robot to make the problem tractable. Deep Reinforcement Learning (DRL) has the potential to enhance the robustness of these algorithms by leveraging our detailed map of the state-action space. This facilitates more efficient exploration and effective policy learning, enabling RL agents to achieve superior performance, quicker convergence, and enhanced generalization to new, unseen situations in real-world applications.

Our main contribution is \nas{} (N-step computation of All Solutions), an algorithm to compute the entire feasible space for a contact planning problem, resulting in a real-time optimal policy for footstep planning. 

Although the theoretical complexity of the algorithm is exponential, we empirically show that it behaves as $O(m*n)$, with $m$ the number of candidate contact surfaces and $n$ the number of steps planned, thanks to the introduction of a node merging procedure that preserves the completeness of our approach.  We also demonstrate the real-time capabilities of the framework to recompute a globally optimal plan.

 In the context of this paper, our results are restricted to gaited locomotion and require to discretise the yaw orientation. Yet the method is not theoretically limited to these use cases. We aim at extending the approach as well as optimising the computation time of the algorithm since it is immediately compatible with GPU parallelisation.

 %In the context of this paper, our results are restricted to gaited locomotion with a fixed yaw orientation, though the orientation can be discretised. We are working on extending \nas{} to address these issues. 

\section{State of the art}

The contact planning problem is a special instance of the motion planning problem, where the objective is to find a collision-free path connecting two configurations of the robot~\cite{piano79}. Planning motions for a legged robot additionally involves planning the discrete change of contacts required to actuate the motion \cite{Bretl2006MotionPO}. This problem is high-dimensional and subject to discrete combinatorics that make it hard to solve.
The central question is to model the contact interactions in a way that reduces the combinatorics.

\subsection{Combinatorics models for gaited locomotion}
Focusing on bipedal gaited locomotion, Chestnutt et al. explicitly deal with the combinatorics by reducing the complexity of the problem through the discretisation of the action space \cite{chestnuttkuffner}, a path further explored recently \cite{griffinastar}. They pre-determine a set of actions that the robot could perform for a fixed position of each end-effector (expressed as target positions for the other end-effector) and use the \astar{} algorithm to plan an optimal path to the goal within this action space.

Deits et al. replaced the discrete action space with a conti\-nuous one, using the classic notion of reachable workspace to linearly constrain the end-effector positions relatively to each other, though the orientation of the robot remains discretised \cite{deits2014footstep}. \astar{} is not immediately applicable with a continuous action space, so Mixed-Integer Programming (MIP) is used to compute a globally optimal solution. MIP has been extended to quadrupeds \cite{aceituno2017mixed, risbourgsolo} and recently the quasi-static constraint on the locomotion \cite{ponton2020efficient, corberes2024perceptive} was overcome. To reduce the computational burden of MIP, we proposed a relaxation of the combinatorics using L1-norm minimisation \cite{tonneau2020sl1m,songslim}, which does not guarantee the convergence to a solution. Other relaxations have since been proposed \cite{MarcucciGCS}. 

Our approach leverages both \astar{} and MIP formulations. We use a dynamic programming approach as in \astar{}, but it is compatible with a continuous action space. Furthermore, we compute the entire feasible space and not a single solution.

\subsection{Combinatorics models for multi-contact locomotion}
Several contributions have also tackled the more general multi-contact problem, where no assumptions are made on the gait followed by the robot and all end-effectors (such as hands) are possibly involved in the contact creations. They are also graph-search methods and have been demonstrated in challenging scenarios, including climbing, chair egress, or tunnel crawling  \cite{escande2009planning,hauser2008motion,wanggraphmc}. As for the gaited locomotion case, the reachable workspace has been used in this context to reduce the dimensionality of the problem and its combinatorics \cite{tonneau2018efficient, tonneau20182pac}. Kumagai et al. \cite{kumagai2020multi} built on this and the notion of contact sustainability \cite{tonneau:tel-01144630, kumagai19} to propose an \astar{} algorithm that efficiently tackles the multi-contact planning problem which we consider to be the state of the art. %The output is a single contact plan, but the formulation is compatible with ours, so extending it to computing all solutions will be the target of our future work.
While we also use dynamic programming, our objective is the computation of all solutions through recursive computation of all reachable states, as similarly done in backward reachability analysis~\cite{bansal2017hamilton,althoff2021set}. For robotics, reachability analysis has been used for motion planning under uncertainty~\cite{lengagne2011planning,liu2017provably,borquez2023hamilton}, while we focus on the contact planning problem.

\subsection{Towards combinatorics-free contact locomotion?}
A variety of approaches have proposed to work around the discrete aspect of combinatorics and relax the contact planning problem into a continuous trajectory optimisation one \cite{mordatch2012DiscCIopt,yunt, Posa-TO-contact, Winkler}. The potential advantage is clear, as the combinatorics is responsible for the exponential complexity of the problem. While these approaches do not guarantee the convergence to feasible solutions, recent work in manipulation suggests that smoothing allows the discovery of solutions to complex scenarios \cite{layeghi2022optimal,pang2023global,le2024leveraging}. 

Deep Reinforcement Learning (DRL) techniques are attractive as they learn a policy efficiently queried online, and have successfully demonstrated their ability to tackle multi-contact locomotion without explicitly modelling the contact decisions \cite{eth19,zhuang2023robot}.  Yet the recent inclusion of model-based optimisation within the training framework has empirically demonstrated the interest in using optimal control with a dynamics model (including contacts) in terms of generalisation and robustness \cite{ethdtc}. A better characterisation of the feasible space for the locomotion problem could alleviate this computational burden and enable the learning of a longer horizon as suggested by \cite{ravi2024efficientsearchlearningagile}.

%the horizon of (about 100 ms in the best scenarios, up to several seconds otherwise), they are not 
%  with results more robust than techniques with an explicit contact model \cite{deepgait}, at the cost of expensive computational time.

%Kufner, Chestnut, Astaratable). Bouyarmane, building on the seminal work of Bretl and Escande and hauser, demonstrated that the \astar{} algorithm can also be applied to multi-contact locomotion. Similarly, Morisawa built on the heuristics proposed by Tonneau to show that planning could be achieved interactively, reducing the computation time from minutes to seconds.

%TODO hauser contact

%% file: definitions.tex
\section{definitions, notations and problem statement}

\newcommand{\pol}[1]{\mathcal{#1}}
\newcommand{\pole}[2]{\mathcal{#1}^{#2}} 
\newcommand{\rkl}[0]{^{r}\mathcal{K}_{l}} 	
\newcommand{\lkr}[0]{^{l}\mathcal{K}_{r}} 	
\newcommand{\ral}[0]{^{r}\mathcal{A}_{l}} 	
\newcommand{\lar}[0]{^{l}\mathcal{A}_{r}} 
\newcommand{\env}[0]{\mathcal{S}} 	
\newcommand{\envj}[1]{\mathcal{S}^{#1}} 	
\newcommand{\tg}[0]{\mathcal{G}} 	

\newcommand{\rotm}{\vc{Q}} 
\newcommand{\fe}[1]{\mathcal{F}_{#1}} 
\newcommand{\re}[1]{\mathcal{R}_{#1}} 
\newcommand{\kee}[0]{^{e}{\mathcal{A}}_{\overline{e}}}
\newcommand{\keem}[0]{^{e}{\mathbf{A}}_{\overline{e}}}
\newcommand{\ree}[0]{^{\vc{p}_e}\mathcal{R}}
\newcommand{\reem}[0]{^{\vc{p}_e}\mathbf{R}}
\newcommand{\pne}[0]{\vc{p}_{\overline{e}}}
\newcommand{\pe}[0]{\vc{p}_{{e}}}
\newcommand{\fer}[0]{^{\pol{R}_e}\mathcal{F}}

\newcommand{\tre}[0]{\mathcal{T}}

\begin{figure}[b!]
     \includegraphics[width=1.0\linewidth]{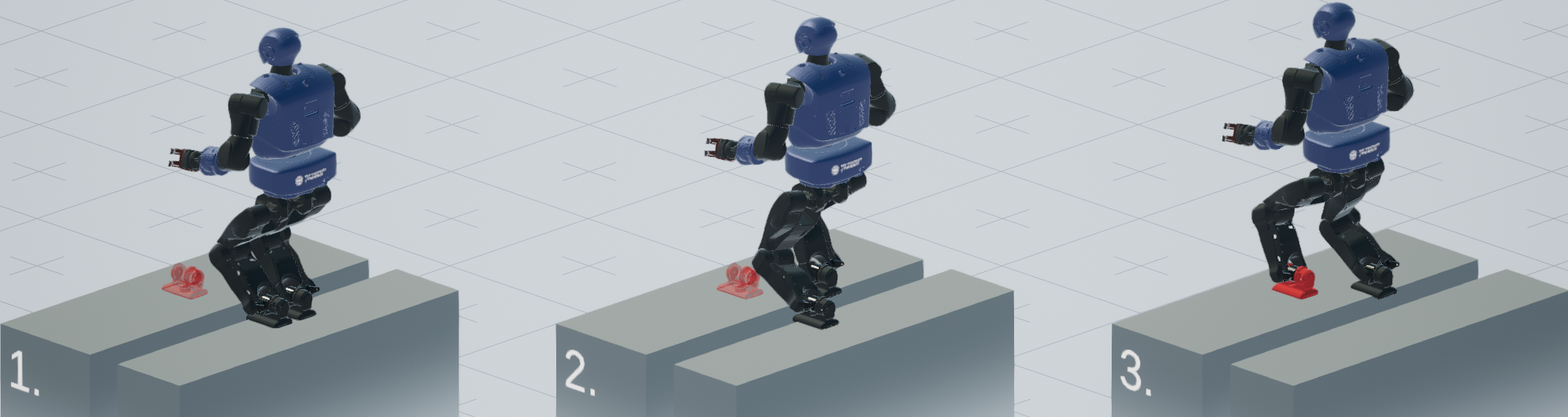}
     \caption{A 2-step plan for the left foot to reach the red target.}
     \label{fig:oversol}
\end{figure}

\begin{figure}[b!]
     \includegraphics[width=1.0\linewidth]{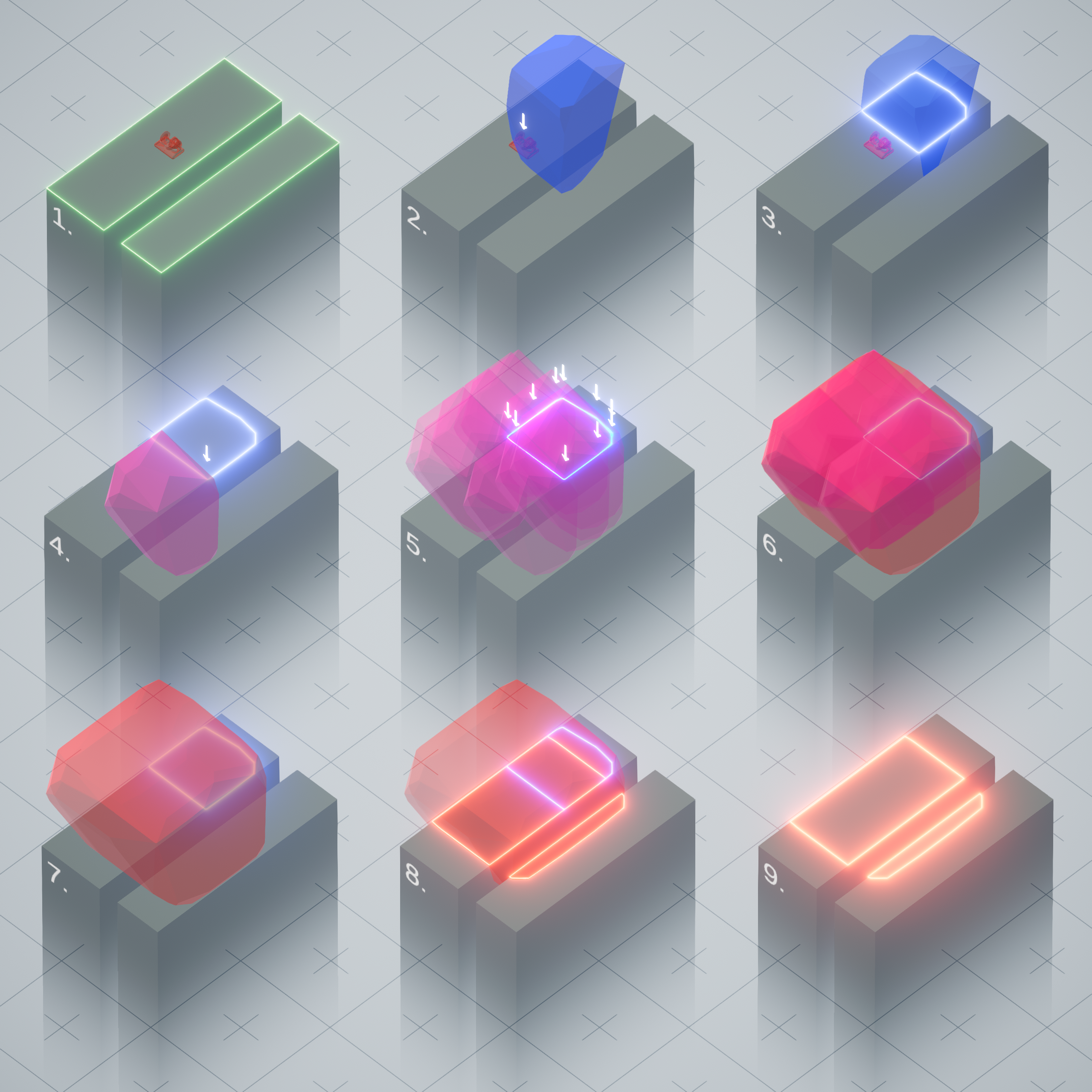}
     \caption{ Two-step feasibility computation. \textbf{1}. Red: The goal for this contact planning problem is to find a feasible contact sequence such that the left foot reaches the red position $\vc{p}_l$, thus we define $\tg = \{\vc{p}_l\}$. Green: The environment $\pol{S}$ is the union of 2 convex polygons. \textbf{2}. All positions of the right foot such that $\vc{p}_l$ can be reached by the left foot in one step are bounded by the blue polytope ${^{\tg}\mathcal{R}}$. It is obtained by translating the antecedent polytope  $\lar$ by $\vc{p}_l$. \textbf{3}. {One}-step feasible set for the right foot ${^{\tg}\mathcal{F}}$. \textbf{4}. Reachability polytope $\ral$ for the position of the right foot indicated by the arrow. \textbf{5}.$\ral$ translated by each extreme point of ${^{\tg}\mathcal{F}}$. \textbf{6}. \& \textbf{7}. Minkowski sum of $\ral$ and  ${^{\tg}\mathcal{F}}$. \textbf{8}. \& \textbf{9}. {Two}-step feasible set for the right foot, composed of 2 convex polygons. }
     \label{fig:overview}
\end{figure}

\subsection{Problem statement}
In the present work, we focus on planning contact sequences for gaited bipedal locomotion tasks. A simple use case for our problem is illustrated in \rfig{fig:oversol}. The figure demonstrates one feasible contact plan that brings the robot from its starting configuration to a goal position, expressed as a terminal constraint on the position of the left foot. Our goal is to capture the infinitely many contact plans that solve this problem.  We assume that the right and left foot alternate in creating contacts towards the goal. For now, we assume that the orientation of the feet around the axis $z$ is fixed for the duration of the planning. The orientation can change along the $x$ and $y$ axes to fit to the contact surfaces.

To simplify the formulation, our equations assume that the left foot always acts as the end-effector reaching the target to complete the task. This assumption is solely for clarity in understanding the algorithm and does not limit our approach.

Given a target goal set $\tg$, the environment, the kinematic constraints of the robot, and a maximum number of steps $n$, our objective is to compute the set of all possible contact sequences that bring the robot to $\tg$ in at most $n$ steps.

\subsection{Surfaces, reachable areas, and kinematic constraints} We use convex polytopes to describe the environment and the reachable workspace of the end-effectors of the robot. The set of feasible solutions to a contact planning problem is described as a union of polytopes.
They are either   3D polyhedra (\rfigtp{fig:overview}{2}) or 2D polygons in a 3D space (\rfigtp{fig:overview}{3}).
A polytope $\mathcal{P}$ is defined as the convex hull of its vertices:
\begin{equation}\label{eq:pol}
\begin{aligned}
    \mathcal{P} := \{\mathbf{p} \in \mathbb{R}^3 | \exists \bm{\lambda} \in \mathbb{R}^{+d},   \|\bm{\lambda}\|_1 = 1 \wedge \mathbf{p} = \mathbf{P} \bm{\lambda} \}  \,.
\end{aligned}
\end{equation}

where $\vc{P} \in \mathbb{R}^{3 \times d}$ is a matrix  obtained by concatenating the $d$ extreme vertices of $\mathcal{P}$ and  $\bm{\lambda} $ is a unit weighting vector of the extreme points. In the following, we implicitly define the matrix $\vc{L}$ for any polytope $\mathcal{L}$ that we define.

\textit{The environment} is represented as a union of $m$ disjoint convex contact surfaces $\env = \bigcup_{j=1}^{m} \envj{j}$ (\rfigtp{fig:overview}{1 - green}).

\textit{The kinematic constraints} of the robot are linearised as commonly done in graph-based approaches. We define  $ \rkl$ as the polytope describing the set of reachable positions for the left foot assuming the right foot is located at the origin. We similarly define $\lkr$ for the right foot. 

\textit{The antecedent constraints} also need to be defined because our algorithm works backward. We define the $\lar$ (\rfigtp{fig:overview}{3 - blue}) as the set of right foot positions from which a left foot position can be placed at the origin of the world. $\ral$ is similarly defined for the left positions such that a right foot position at the origin is reachable (\rfigtp{fig:overview}{4 - red}).  $\ral$ and $\lar$ are obtained by applying a central symmetry to the vertices of $\rkl$ and $\lkr$.

\textit{The goal} $\tg$ is the polytope defining the task for a robot. The plan is successful when either end-effector reaches $\tg$. $\tg$ is a subset of a contact surface $\mathcal{S}^j$, and can be a degenerated polytope (i.e. a single point, as in \rfigtp{fig:overview}{1 - red}).

%% file: algo.tex
\section{Overview}
We propose to compute the set of feasible contact sequences to $\tg$ using a dynamic programming algorithm. Starting from $n=0$, we recursively compute all feasible states from which we can reach $\tg$ in $n$ steps, until a termination condition is met, either if the current state of the robot is reached or a user-defined $n$ is reached.

By definition, the feasible set  $\fe{0}$ for $n=0$ (\emph{i.e.} the target is reached without making any steps) is $\tg$ (\rfigtp{fig:overview}{1}). We then compute the reachable set $\re{1}$  of all positions that can bring the end-effector to $\tg$ in one step, for $n=1$ (\rfigtp{fig:overview}{2}).  

Most of the positions in $\re{1}$ are not contact positions, so to compute the feasible set 
$\fe{1}$ we intersect $\re{1}$ with $\pol{S}$, giving $\fe{1} = \bigcup_{j=1}^{m} \fe{1}^j$, with $\fe{1}^j = \re{1} \cap  \envj{i}$ (\rfigtp{fig:overview}{3}).

Each non-empty $\fe{1}^j$ describes a 2D polygon such that for any position of the right foot on the polygon, there exists a one-step sequence that results in the left foot inside $\tg$. Each $\fe{1}^j$ corresponds to a node added as child to $\fe{0}$ in a tree $\tre$. 

We can then proceed similarly to compute the feasibility set $\fe{2}^j$ associated with each $\fe{1}^j$ (\rfigtp{fig:overview}{4-9}), and recursively compute the feasibility sets until we reach $\fe{n}$.

From any node of $\tre$, the contact sequence to $\tg$ is obtained by recursively selecting the parent node until $\tg$ is reached.

At runtime, to compute a contact sequence from a given state of the robot, we use a \kdt{} to efficiently search for the node corresponding to the state of the robot.

\section{1-step feasibility}

Similarly to other Dynamic Programming algorithms, we reason from the goal $\tg$. The \textit{antecedent} states form the set of all positions of an end-effector from which $\tg$ can be reached in exactly one step. We rely on \rfigtp{fig:overview}{} to illustrate the procedure. It describes the computation of the 2-step feasible set for the problem in \rfig{fig:oversol} and contains all the possible cases that can occur when computing the \textit{antecedent} states.

\subsection{Reachability from a given position (\rfigtp{fig:overview}{1-2})}

For a given position $\vc{p}_e$ of an end-effector $e$ on flat ground, the set of positions of the other effector $\overline{e}$ such that $\vc{p}_e$ is reachable in one step is by definition $\kee$ translated by $\pe$:

\begin{equation*}
\begin{aligned}
    \ree:= \{&\pne \in \mathbb{R}^3 | \exists \bm{\lambda} \in \mathbb{R}^{+d}, \\ &|| \bm{\lambda}|| = 1 \wedge \pne = \keem \bm{\lambda} + \mathbf{p_e} = \reem \bm{\lambda} \}  \,.
\end{aligned}
\end{equation*}

 In the general case, if $\rotm$ is the rotation matrix of minimum distance (in the log sense) that aligns the robot's root frame $z$ axis with the contact surface normal, we write:

\begin{equation*}
\begin{aligned}
    \ree:= \{&\pne \in \mathbb{R}^3 | \exists \bm{\lambda} \in \mathbb{R}^{+d}, \\ & || \bm{\lambda}|| = 1 \wedge \pne = \rotm \keem \bm{\lambda} + \mathbf{p_e} = \reem \bm{\lambda} \}  \,.
\end{aligned}
\end{equation*}

\newcommand{\rer}[0]{^{\pol{R}_e}\mathcal{R}}
\newcommand{\reb}[0]{\pol{R}_e}
\newcommand{\rem}[0]{\vc{R}_e}
\newcommand{\rerm}[0]{^{\pol{R}_e}\mathhbb{R}}

\subsection{Reachability from a set of positions (\rfigtp{fig:overview}{4-7})}
More interestingly, we can compute the antecedent set of a convex polytope $\pol{R}_e$ of positions of ${e}$ that share the same contact normal. This set  $\rer$ is all the positions for $\overline{e}$  allowing to create a contact with $e$ inside $\pol{R}_e$ in one step. It is thus  the union of all antecedents at every point of $\pol{R}_e$:
\begin{equation}\label{eq:rer}
\begin{aligned}
    \rer:= \{ \rotm \keem \bm{\lambda} +    \rem\bm{\lambda}_1, \ ||\bm{\lambda}|| = ||\bm{\lambda}_1|| = 1 \}  \,.
\end{aligned}
\end{equation}

With $\bm{\lambda}$ and $\bm{\lambda}_1$ positive vectors of appropriate dimensions. 
Eq(\ref{eq:rer}) denotes the Minkowski sum of the two convex sets $\pol{R}_e$ and $\kee$ (the latter being rotated by $\vc{Q}$). Therefore, $\rer$ is convex since the Minkowski sum preserves convexity. 

%If  $ ^{\pol{R}_e}\mathcal{R}$ is a convex polygon (as is always the case in this work), using a local frame aligned with $ ^{\pol{R}_e}\mathcal{R}$, the Minkowski sum can be computed in linear time with the number of vertices~\cite{berg2008} \stn{(TODO Demonstarte for 3D )}.

\subsection{Computing the 1-step feasible set (\rfigtp{fig:overview}{7-9})}
$\fer$ is the subset of $\rer$ that results in contact locations. $\fer$ is obtained by intersecting $\rer$ with the contact surfaces of the environment and, whenever a collision is detected, computing the resulting intersected surfaces. $\fer$ is thus composed of the union of all resulting surfaces:

$\fer= \bigcup_{j=1}^{m} {\fer^j}$,  with $\fer^j = \rer \cap  \envj{i}$.

If the set of positions given as input is a convex polytope (as is the case when we start from $\tg$), all resulting surfaces are convex polygons, since the intersection of 2 convex polytopes is a convex polytope. This means that at any point of the expansion, \textbf{the feasible set is always a union of convex polytopes} (possibly degenerated into a point). % This means that we now have all the tools needed to compute the feasible set for any number of steps.

\section{The \nas{} algorithm}
To compute the N-Step feasible space, that is the set of all possible positions from which the robot can reach $\tg$ in at most $n$ steps, we recursively compute the feasible set for all the steps from 1 to $n$ in order to populate a tree $\tre$.

\subsection{Tree description and initialisation}

Each node of $\tre{}$ contains information about the end-effector currently in contact, the surface in contact, as well as the subset of the surface covered by this node. It also contains a link to its parent in the $\tre$ as shown in \rstruct{alg:strcut}.

\captionsetup[algorithm]{labelformat=empty}
\begin{algorithm}
\caption{\textbf{struct 1} \textsc{Node}}
\begin{algorithmic}%[1] % uncomment for line numbers
%\Struct{Node}
%\State \textbf{struct} \textsc{Node}
  \State \hspace{1em} $effectorId$ : \textsc{enum} 
  \State \hspace{1em} $parent$ : \textsc{Node*} // parent node
  \State \hspace{1em} $surfaceId$ : \textsc{int}
  \State \hspace{1em} $extremePoints$ : \textsc{Point list} //polygon description
 % \EndStruct
\end{algorithmic}
 \label{alg:strcut}
\end{algorithm}

\captionsetup[algorithm]{labelformat=default}

$\tre$ is implemented as an array of Node lists. Nodes are indexed by their depth in the tree, which is the number of steps  required to reach the target from the node. % $\tre$  is initialised with a single node (at depth 0) corresponding to $\tg$. For simplicity in this paper, we again consider that for this root node, the $effectorId$ matches the left foot.

\subsection{The \nas{} algorithm}

\begin{algorithm}
\caption{\nas}\label{alg:compute_feasibility}
\begin{algorithmic}
\Function{N\_STEPS\_FEASIBILITY}{$\tre$, $n$}
 
    \If{$n == 0$} \textbf{ \Return $\tre$ // (\rfigtp{fig:overview}{1})}
\EndIf
    \For{ each leaf node $node$ in $\tre$}
        \State $feasibleNodes \gets$ \Call{FEASIBLE\_NODES}{$node$}
        \State \Call{add\_leaves}{$\tre$, $node$,  $feasibleNodes$}
    \EndFor    
    \State \Return \Call{N\_STEPS\_FEASIBILITY}{$\tre$, $n-1$} 
\EndFunction \\
\end{algorithmic}

\begin{algorithmic}
\Function{FEASIBLE\_NODES}{$node$}
    \State $effId \gets \Call{other}{node.effectorId}$\State $nodeLists \gets []$ 
    \State $\pol{R} \gets \Call{REACH\_POLYTOPE}{node}$ //  (\rfigtp{fig:overview}{2})
    \For{each surface $s$ in $\pol{S}$}
        \State $feasible\_s \gets \Call{Intersect}{\pol{R}, s}$ //  (\rfigtp{fig:overview}{3,8})
        \If {\Call{not\_empty}{$feasible\_s$}}
            \State \footnotesize{$child \gets \Call{Node}{effId, node, s.Id, feasible\_s}$}
            \State \normalsize{$nodeLists.add(child)$}
        \EndIf
    \EndFor
    \State \Return $nodeLists$
\EndFunction
\end{algorithmic}
\end{algorithm}

\nas{} consists in initialising $\tre$ with a single node $\tg$ before calling N\_STEPS\_FEASIBILITY (\ralg{alg:compute_feasibility}). 
N\_STEPS\_FEASIBILITY is a recursive function that computes the expansion of  $\tre$ for 1 step before calling itself. Each call computes the 1-step feasible set associated with each leaf node (method FEASIBLE NODES), creates children nodes, and adds them to the leaf node (method ADD\_LEAVES).

FEASIBLE\_NODES first computes which end-effector is selected for the expansion (method OTHER). The method REACH\_POLYTOPE computes the volume $\pol{R}$ from which the node can be reached as per eq.(\ref{eq:rer}). 
$\pol{R}$ is then intersected with each potential contact surface from $\pol{S}$ (method INTERSECT). Each non-empty intersection results in a new node.

\subsection{Discrete handling of the rotation}
\nas{} can be extended to handle a rotation of the foot around the $z$ axis, if we  discretise the possible orientations~\cite{deits2014footstep}. The updated function FEASIBLE\_NODES is given by \ralg{alg:feas2}. 
ROTATE rotates $\pol{R}$ around the $z$ axis. Additionally, the current rotation angle of the feet needs to be added to the node structure to know the end-effector orien\-tation at the current state. This algorithm is presented here for completeness, but our experiments focus on \ralg{alg:compute_feasibility}. 

\begin{algorithm}
\caption{FEASIBLE\_NODES with yaw orientation}
\label{alg:feas2}
\begin{algorithmic}
\Function{FEASIBLE\_NODES}{$node$}
    \State $effId \gets \Call{other}{node.effectorId}$\State $nodeLists \gets []$ 
    \State $\pol{R} \gets \Call{REACH\_POLYTOPE}{node}$  
    \color{red}\For{each discrete angle value $\theta$}
    \State $\pol{R}_{\theta} \gets \Call{Rotate}
    {\pol{R}, \theta}$
    \color{black}
    \For{each surface $s$ in $\pol{S}$}
        \State $feasible\_s \gets \Call{Intersect}{\color{red}{\pol{R}_{\theta}}, s}$
         \If {\Call{not\_empty}{$feasible\_s$}}
            \State { \color{red}{$\gamma \gets  node.angle + \theta$}}
            \State \footnotesize{ $child \gets \Call{Node}{effId, node, s.Id, feasible\_s, \color{red}{\gamma}}$}
            \State \normalsize{$nodeLists.add(child)$}
        \EndIf
    \EndFor
    \EndFor
    \State \Return $nodeLists$
\EndFunction
\end{algorithmic}
\end{algorithm}

\subsection{Properties of $\tre$ }
%The following properties immediately results from $\nas{}$:

%\subsubsection{Linear time Contact surface sequence computation}
%A valid contact surface sequence is a list of contact surfaces such a valid contact sequence can be found if steps are sequentially taken on each surface. From   from any node to $\tg$ it is immediately given by selecting the polygon associated to the node, and repeating the process with the node's parent, until $\tg$ is reached. This process is thus linear with the number of steps in the sequence, ie $O(n)$.

\subsubsection{n-step completeness}
For a given robot state, defined as the position $\vc{p}_e$ of the active end-effector on a contact surface, if there exists a contact sequence leading to the target in $n$ steps, there is necessarily a node in $\tre$ that contains $\vc{p}_e$\footnote{We abusively refer to a node containing a point or a state to indicate that the polygon associated with the node contains the point of interest.}. Otherwise, 
there is no valid up-to-$n$-steps contact sequence.

\subsubsection{Minimum step optimality}
For a given $\vc{p}_e$, there can be more than one matching node. The nodes with the lowest depth all denote a sequence with a minimum number of steps.

\newcommand{\olh}[0]{O(\log(h)}
\newcommand{\oln}[0]{O(n)}
\newcommand{\olhn}[0]{O(n+\log(h))}

\subsection{Optimisation of the algorithm}
\label{sec:optim}

\subsubsection{Optimising the tree generation}
Unsurprisingly, \nas{} has a theoretical exponential complexity in $O(b^n)$, with $b \leq m$  the branching factor (or average number of successors per node). This is aligned with the worst case \astar{} and mixed-integer complexities. However, this complexity can be reduced through the use of what we call node merging.

The expansion of several leaf nodes at step $j$ can lead to new leaf nodes at step $j+1$ covering exactly the same surface. This can happen when all points reachable on a given surface have been covered by the expansion. It is possible to reduce the number of branches by merging such nodes into a single one with several parents, without changing the completeness nor the optimally properties of $\tre$: for bidepal locomotion the position of the parent end-effector has no influence on the expansion of the node, and since we keep track of all parents no path is lost. This optimisation can also be applied if more than 2 end-effectors are involved but it requires considering all end-effectors when merging.

%Another optimisation consists in detecting ``cycling expansions''. If a node covers the exact same surface as its grandparent it means that the expansion from this node will only reach points visited more optimally from the grandparent (\emph{i.e} 2 step shorter). If exhaustivity is not required these nodes can be removed from the tree as they can only add extra steps ``in place'' to existing contact plans. %\stto{justify if we dont have the time to compute numbers for this by saying we want to be exhaustive}
%If the surface of one node is completely included in the surface another node, it is also possible to merge the nodes but this either constrains to plan the footstep sequence with a horizon of one (or more if there is a chain of parent merged nodes), or to store both surfaces within the node adequately. In any case, stepping anywhere on the largest surface will always result in an optimal contact plan. Our results show that node merging significantly reduce the total number of nodes in $\tre$. 
%(\stto{NUMBERS+ clarify})
%(\stto{Include surface cycles ?})

\subsubsection{Optimising the tree exploitation using a kd-tree}
The algorithms used for contact planning will require identifying which node(s) of  $\tre$ ``contain(s)'' a given point $\vc{p}_e$, i.e. nodes such that $\vc{p}_e$ is included in their associated polygon. Finding such a node involves iterating through all the nodes sorted by their depth in $\tre $ until encountering one that contains $\vc{p}_e$ (it will be an optimal one). This process has a linear complexity.
To improve efficiency, we store the nodes in a separate \kdt{} \cite{kdtree} for each end-effector. This allows for an average search complexity of $\log(h)$, with $h$ the total number of nodes. The construction complexity of the \kdt{}, in $O(h\log(h)$, is dominated by the complexity of constructing $\tre$. Additionally, the \kdt{} can efficiently return all the nodes containing  $\vc{p}_e$.

%\subsubsection{Cycling expansion}
%If a node covers the exact same surface as its grandparent it means that the expansion from this node will only reach points visited more optimally from the grandparent (\emph{i.e} 2 step shorter). If exhaustivity is not required these nodes can be removed from the tree.
%\subsubsection{Avoiding ``surface cycles''}
%As we are only dealing with convex surfaces, there are no real practical reasons for the same effector to leave a contact surface before returning on the same surface 2 steps later. For that reason, nodes that introduce such redundancy are not added to the graph. This is easily achieved by considering the surfaces contacted by the same effector 2 and 4 steps before the current step. This optimisation should not be applied if exhaustivity is required.
\subsection{Applications for  \nas{}}
Most of the following use cases rely on the search of a specific node in $\tre$. The resulting complexities indicated assume that the search is implemented using the \kdt{} for a complexity in $O(\log(h))$, with $h$ the number of nodes in $\tre{}$, as detailed in \rseq{sec:optim}.

\subsubsection{$\tre$ as an optimal policy for contact planning}
For a given state of the robot $\vc{p}_e$, we can find any node that contains $\vc{p}_e$ (and the associated $effectorId$), then go up the parent node chain to compute a contact sequence to the target. Assuming the depth of the node is $k \leq n$, this results in a feasible contact plan $\vc{F}=[\pol{F}_{k-1,e},\cdots,\pol{F}_{1,e},\pol{F}_{0,e} \subset \tg{}]$,  where $\pol{F}_{i,e}$ is the feasible convex polygon given by the node at depth $i$ in the selected sequence. This process has an average complexity of $\olhn$.

To find a contact sequence with a minimal number of steps, we can select the first node at the lowest depth which contains $\vc{p}_e$. %This is done at no extra cost if the nodes are sorted by depth in the kd-tree.
Alternatively, if $\vc{p}_e$ is known at expansion time, the algorithm terminating condition can be modified to stop the expansion whenever a node containing $\vc{p}_e$ is generated (or until a maximum iteration is reached, meaning that the problem has no solution). This node will always result in a path that involves the minimum number of steps to the target.

\subsubsection{Computing exact footstep locations}
$\tre$ is a policy for selecting contact surfaces, but does not directly tell us where exactly on the surface the stepping should occur. By construction of the graph, any point included by a node is optimal regarding the minimum number of steps. This allows us to apply any selection technique for a valid contact position, as long as we choose a point reachable from the previous location of the robot. The closest reachable point from the Chebyshev center of the node can be selected for robustness for instance. 
If a given objective has to be minimised for the footstep plan, a linearly constrained optimisation problem can be solved. The user is free to consider as many steps as needed in the optimisation horizon $n_{hor}$: no matter  $n_{hor}$, the positions selected will always be feasible and lead to a minimal number of steps. From a current position $\vc{p}_0$ within a node at depth $k > 0$, an example of convex program is:

\begin{equation}\label{eq:qp}
\begin{aligned}
    \textbf{find} \quad & \mathbf{X} = [\mathbf{p}_1,\cdots,\mathbf{p}_{n_{hor}}] \in \mathbb{R}^{3\times n_{hor}} \\ 
    \textbf{min} \quad & l(\mathbf{X}) \\
    \textbf{s.t.}
    \quad & \forall i, 1 \leq i \leq n_{hor} :& \\ 
    \quad & \quad \quad \mathbf{p}_{i} \in  {^{\vc{p}_{i-1}}\mathcal{R}} \cap \pol{F}_{k-i,0} 
\end{aligned}
\end{equation}

With $l$ a convex objective, $1 \leq n_{hor}\leq k$ and ${^{\vc{p}_i}\mathcal{R}}$ the reachable workspace from the contact $\vc{p}_i$.

\subsubsection{Online replanning using $\tre$}
At any point during the motion of the robot, the contact plan can be efficiently updated if the situation invalidates it. Whenever the robot makes a new step, we can check whether the contact location $\vc{p}_i$ belongs to the planned $\pol{F}_{i-1,e}$. If not (e.g. due to a perturbation of the hardware) we can immediately query $\tre$ again and obtain an updated path from $\vc{p}_i$.

The formulation also allows to dynamically mark areas of the scene as impassible. To remove a contact surface from the feasible set, we mark all the nodes concerned by the surface as invalid. Upon re-planning, from the current state of the robot, we can iterate through the nodes found until we find a sequence that does not go through any invalid nodes. The average complexity for the search is lower than the search for the nodes containing the current position ($O(\log(h))$), plus at most $v$ times the exploration of paths each of length of at most $n$, giving $O(n*v + \log(h))$, where $v$ is the number of nodes that contain the current robot state.

\subsubsection{Optimal trajectory optimisation}
\nas{} can also be formulated as a single-query \astar{} algorithm with continuous nodes instead of discrete points, with an expansion that is not exhaustive but rather guided by a heuristic, for a likely better average complexity. The main advantage over a standard formulation is that the formulation is continuous and reduces the branching factor. We leave the comparison of both approaches for future work as the focus of this paper is on the characterisation of all solutions.

\subsection{Implementation details}
We implement the Minkowski sum of a polygon $\pol{S}$ and a polyhedron $\pol{P}$ as the convex hull of the polytopes obtained by translating $\pol{P}$ by each vertex of $\pol{S}$. The complexity of this operation is $O(k \log (k))$ where $k = k_{\pol{S}} *k_{\pol{P}}$ and $k_{\pol{S}}$ and $k_{\pol{P}} $ are the total number of vertices in $\pol{S}$ and $\pol{P}$. This operation remains efficient as $k_{\pol{S}}$  typically remains below 10. 
%By adapting \cite{berg2008} it is possible to write an algorithm that has a linear complexity, however in our 
We compute the convex hulls using the SciPy ~\cite{2020SciPy-NMeth}.
Our code is implemented in Python, including collision detection and the \kdt, with efficiency not being the primary concern. %As such, the computation times reported are only indicative and could probably be largely improved, although they are already compatible with real time applications.

To avoid computing plans that result in the foot partially out of a contact surface, we automatically scale down the surfaces given as input to \nas{}. This parametrisation is optional if the case is handled by the controller as in~\cite{griffin2019footstep}.

%% file: experiments.tex
\section{experiments}

We report quantitative information about the generation and run time of \nas{} that empirically demonstrate the vali\-dity of the approach on the Talos robot \cite{stasse2017talos}. The experiments were run on an desktop computer equipped with an Intel i9 9900K CPU (3.6GHz) and 64GB RAM. 
The motions were synthesised using the PAL robotics controller. They are validated on the real robot or on the Gazebo simulator \cite{gazebo}, synchronised in real-time with Unity 3D for rendering~\cite{unity3d}. 

We designed 5 different scenes, 4 of which are shown in the teaser figure, with a number of surfaces varying from 4 to 43. One scene includes non-flat surfaces. Our video (\url{https://youtu.be/I5yFe0ez0sI}) demonstrates examples of minimum step motions computed with  \nas{}, including an example of dynamic re-planning. % To better reflect the capabilities of the framework, our tests are conducted by only considering ``complete'' formulations $\nas{}$, which means that node merging is the only allowed optimisation. We also assume a fixed yaw orientation.

\subsection{Tree generation analysis}
\begin{figure}[htbp]
\centering
        \includegraphics[width=0.7\linewidth]{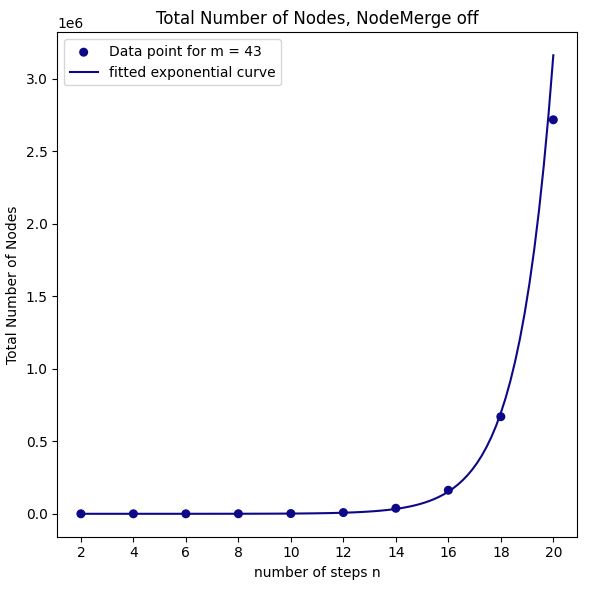}
    \caption{Exponential growth of nodes without node merging.}
    \label{fig:nomerge}
\end{figure}

\begin{figure}[htbp]
    \centering
    \begin{subfigure}[b]{0.8\linewidth}
        \centering
        \includegraphics[width=\linewidth]{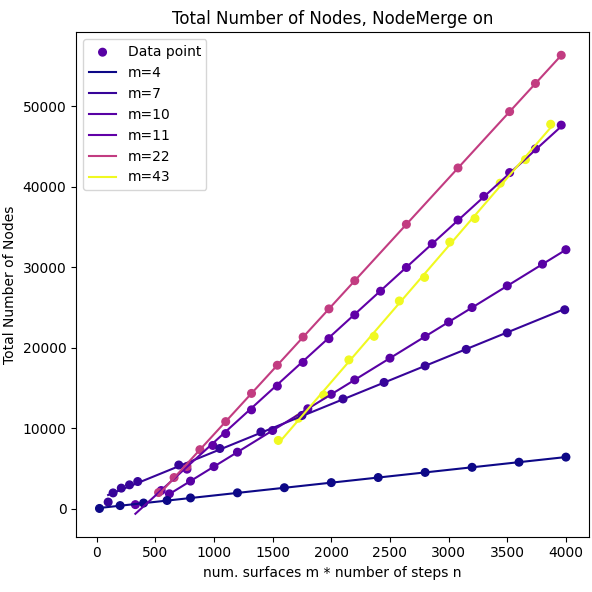}
    \end{subfigure}
    \vfill
    \begin{subfigure}[b]{0.8\linewidth}
        \centering
        \includegraphics[width=\linewidth]{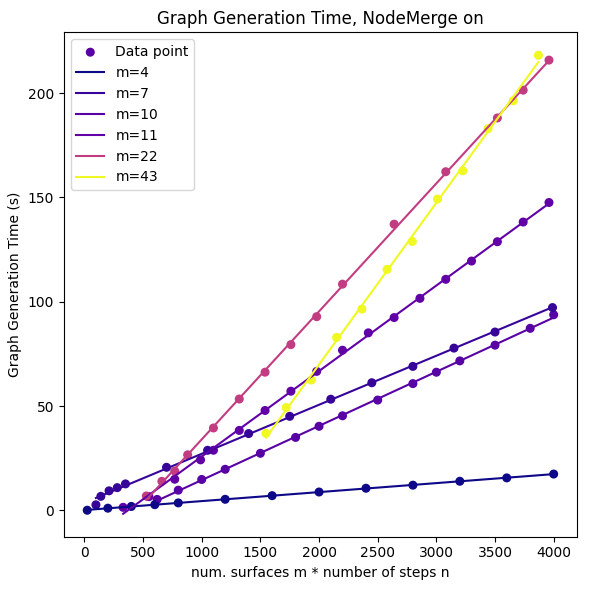}
    \end{subfigure}
    \caption{In all instances, node merging allows the number of nodes and generation time for $\tre{}$ to grow as $O(m*n)$.}
    \label{fig:merge}
\end{figure}

The theoretical complexity of the graph generation is $O(b^n), b \leq m$. This is confirmed by \rfig{fig:nomerge}, which presents the number of nodes in $\tre{}$ when no optimisation is used in one representative use case. However, node merging optimisation results in a bilinear complexity $O(m*n)$ in all scenarios, as evidenced by \rfig{fig:merge}.  Even in unrealistic scenarios (such as planning 100 steps over a scene with more than 40 surfaces), the graph generation time remains below 5 minutes. This proves empirically the viability of \nas{}.

\subsection{Tree exploitation}
For lack of space, we report briefly on the exploitation of the tree. The \kdt{} query time is conditioned by the number of nodes, which in all the scenarios demonstrated remains below 10,000. In this context, the query time is below 24 ms in our scenarios and at worst 89 ms (for the scene with 43 surfaces), within our 10Hz requirement.

To compute the footstep sequences we solve \req{eq:qp} with the complete horizon (although we have established that this is not a requirement) and $l=0$\footnote{Hence we only solve a feasibility problem.}. In all instances, the cumulated time to query the \kdt{} and solve the QP remains below 100 ms. Here the resolution time only depends on $n$ as the combinatorics is fixed: for instance, the 43 surfaces scene is solved in $6$ ms ($n = 23$),  while the re-planning scene is the longest to solve with $79$ ms (and $n=61$).

\section{discussion}
Our experiments show that \nas{} is applicable to typical planning problems for biped robots. The main benefit of the formulation is that it comes with strong guarantees, within the assumptions that are made when defining the problem.

\subsubsection{Handling multiple goals}
As a dynamic programming approach \nas{} needs to be computed from the goal for the feasibility to be guaranteed. Expanding $\tre$ from the starting position would allow to tackle multiple goals, but online-replanning would not be possible past the first step.

\subsubsection{Computing n}
The number of steps $n$ can be initially automatically computed, as we stop the expansion as soon as the current robot position is covered by $\tre$. However, if re-planning involves that the new optimal motion requires more than $n$ steps, a new expansion phase needs to be recomputed online until the new value of $n$ (again, optimal) is found.

\subsubsection{Handling continuous yaw orientation}
Our approach (and the state of the art) currently only allows handling the yaw orientation of the end-effectors in a discretised manner. We are working on a continuous formulation that involves extending the reachability formulation to four dimensions.

\subsubsection{Scaling the approach to non-gaited loco-manipulation}
\nas{} can be directly extended to any legged robots. However, the size of $\tre{}$ will grow significantly as the result of introdu\-cing additional discrete choice, in particular if locomotion is not gaited, although node merging remains possible. Future work will establish under what conditions this extension is viable, specifically if the computation of $\tre{}$ can be optimised. 

\subsubsection{Parallelisation}
As a breadth-first algorithm, \nas{} is parallelisable, which could  significantly improve the computation times. Furthermore, the operations involved in the expansion are all compatible with a GPU implementation, which could further improve this efficiency.

\subsubsection{Interest for machine learning}
We argue that the cha\-racterisation of the complete feasible space, even under simplifying assumptions, presents two advantages for learning:
\begin{itemize}
    \item The tailoring of the search space to the close neighborhood of the feasible set, to improve sample efficiency;
    \item The complete combinatorics can be explored and relevant information regarding optimality and whole-body feasibility can be fed back to the training network.
    
\end{itemize}

\section{Conclusion}
In this work, we present \nas{}, a dynamic programming algorithm for computing the feasible space of a contact planning problem. \nas{} computes a globally optimal policy for a given problem, which allows for real-time planning (and re-planning) of a feasible contact plan. Thanks to the node merging procedure we introduce, the computation of the feasible space is performed with a bilinear complexity. 

The reachability-based formulation of \nas{} also enables the implementation of a novel, continuous \astar{} algorithm to solve the problem once, with optimality guarantees equivalent to recent Mixed-Integer formulations.

\nas{} is parallelisable and could be considered for impro\-ving the efficiency of machine learning algorithms thanks to the tight characterisation of the search-space for the problem.

\section{Acknowledgements}
The authors would like to thank Rajesh Subburaman for his help on hardware experiments.